\documentclass[11pt,a4paper]{article}
\usepackage[hyperref]{eacl2021}
\usepackage{times}
\usepackage{latexsym}

\usepackage{graphicx}
\usepackage{makecell,booktabs}
\usepackage{enumitem}
\usepackage{float}
\restylefloat{table}
\usepackage{subcaption}
\usepackage{microtype}
\usepackage{subcaption}
\usepackage{amsmath}
\usepackage{subcaption}

\usepackage{mathtools}

\usepackage{microtype}

\aclfinalcopy %

\newcommand\cmu{$^1$}
\newcommand\technion{$^2$}
\newcommand\aspace{\hspace{.75em}}

\title{Probing the Probing Paradigm: \\ Does Probing Accuracy Entail Task Relevance?}

\author{
 Abhilasha Ravichander\cmu \aspace \aspace
 Yonatan Belinkov$^2$\thanks{~~Supported by the Viterbi Fellowship in the Center for Computer Engineering at the Technion.}  \aspace \aspace
 Eduard Hovy\cmu \\
 \cmu Language Technologies Institute, Carnegie Mellon University \\
 \technion Technion -- Israel Institute of Technology \\ 

  {\tt aravicha@cs.cmu.edu} \\
  {\tt belinkov@technion.ac.il}, \hspace{1em} {\tt hovy@cmu.edu} 
}

\date{}

\begin{document}
\maketitle
\begin{abstract}
Although neural models have achieved impressive results on several NLP benchmarks, little is understood about the mechanisms they use to perform language tasks. Thus, much recent attention has been devoted to analyzing the sentence representations learned by neural encoders, through the lens of `probing' tasks. However, to what extent was the information encoded in sentence representations, as discovered through a probe, actually used by the model to perform its task? In this work, we examine this probing paradigm through a case study in Natural Language Inference, showing that models can learn to encode linguistic properties even if they are not needed for the task on which the model was trained. We further identify that pretrained word embeddings play a considerable role in encoding these properties rather than the training task itself, highlighting the importance of careful controls when designing probing experiments. Finally, through a set of controlled synthetic tasks, we demonstrate models can encode these properties considerably above chance-level \emph{even when distributed in the data as random noise}, calling into question the interpretation of absolute claims on probing tasks.\footnote{Code and data available at \url{https://github.com/AbhilashaRavichander/probing-probing}.}
\end{abstract}
\section{Introduction}
 
Neural models have established state-of-the-art performance on several NLP benchmarks \cite{kim-2014-convolutional, DBLP:conf/iclr/SeoKFH17, chen-etal-2017-enhanced,  devlin-etal-2019-bert}. However, these models can be opaque and difficult to interpret, posing barriers to widespread adoption and deployment in safety-critical or user-facing settings \cite{belinkov2019analysis}. How can we know what information, if any, neural models learn and leverage to perform a task?  This question has spurred considerable community effort to develop methods to analyze neural models, motivated by interest not just to have models perform tasks well, but also to understand the mechanisms by which they operate.

\begin{figure}[!tb]

    \includegraphics[width=\columnwidth]{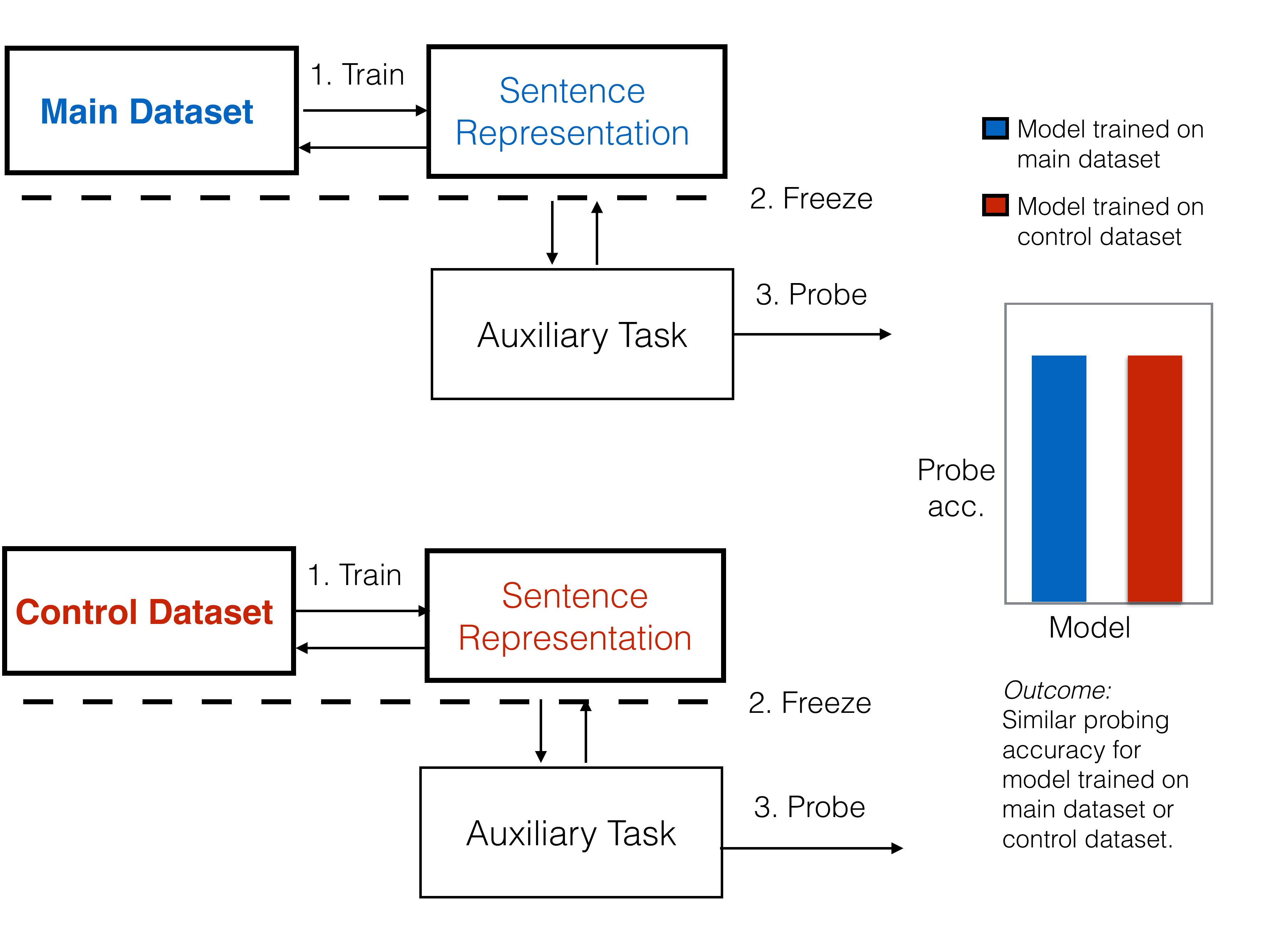}
  \caption{Illustration of our control dataset methodology for evaluating probing classifiers. Control datasets are constructed such that a linguistic feature is not discriminative with respect to the task. Representations from models trained on the main dataset and control dataset are probed for the linguistic feature, and demonstrate similiar probing performance. }
 \label{ref:probing-exp}
\end{figure}
A popular approach to model introspection is to associate the representations learned by the neural network with linguistic properties of interest, and examine the extent to which these %
properties can be recovered from the representation \cite{adi2017finegrained}. This paradigm has alternatively been called probing \cite{conneau-etal-2018-cram}, auxiliary prediction tasks \cite{adi2017finegrained}  and diagnostic classification \cite{veldhoen2016diagnostic, hupkes2018visualisation}.  As an example of this approach, let us walk through an application to analyze information about tense stored in a Natural Language Inference (NLI) model. In \citet{conneau-etal-2018-cram}, three sentence-encoder models are trained on a NLI dataset \citep[MultiNLI;][]{williams-etal-2018-broad}. The encoder weights are frozen, and the encoders are then used to form sentence representations for the auxiliary task--- predicting the tense of the verb in the main clause of the sentence. A separate classifier, henceforth called the probing classifier, is trained to predict this property based on the constructed representation. The probing task itself is typically selected to be relevant to the training task, and high probing performance is considered as evidence that the property is encoded in the learned representation. Due to its simplicity, a growing body of work uses this approach to pinpoint the information models rely on to do a task \cite{alt-etal-2020-probing, giulianelli2018hood, saleh-etal-2020-probing}.

In this work, we examine the connection between the information encoded in a representation and the information a model relies on. Through a set of carefully designed experiments on the benchmark SentEval probing framework \cite{conneau-etal-2018-cram}, we shed light on information use in neural models. Our story unfolds in four parts:
\begin{enumerate}
\item First, we establish careful control versions of the training task such that task performance is invariant to a chosen linguistic property (Figure~\ref{ref:probing-exp}). We show that even when models cannot use a linguistic property to perform the task, the property can be reliably recovered from the neural representations through probing (\S\ref{ref:incidentalNLI}).
\item Word embeddings could be a natural suspect for this discrepancy. We demonstrate that initializing models with pretrained word embeddings does play a role in encoding some linguistic properties in sentence representations. We speculate that probing experiments with pretrained word embeddings conflate two tasks --- training word embeddings and the main task under consideration (\S\ref{ref:effectwordembedding}). 
\item What happens if we neutralize the effect of pre-trained word embeddings? Even when word embeddings are trained from scratch, we demonstrate that models still encode linguistic properties when they are not actually required for a task (\S\ref{ref:isolae}).
\item Finally, through a carefully controlled synthetic scenario we demonstrate that neural models can encode information incidentally, even if it is distributed as random noise with respect to the training task (\S\ref{ref:adversarial}). We discuss several considerations when interpreting the results of probing experiments and highlight avenues for future research needed in this important area of understanding models, tasks and datasets (\S\ref{ref:discussion}).
\end{enumerate}

\section{Background and Related Work}
\label{ref:relatedwork}

Progress in Natural Language Understanding (NLU) has been driven by a history of defining tasks and corresponding benchmarks for the community \cite{marcus-etal-1993-building, dagan2006pascal, rajpurkar-etal-2016-squad}. These tasks are often tied to specific practical applications, or to developing models demonstrating competencies that transfer across applications. The corresponding benchmark datasets are utilized as proxies for the tasks themselves.  How can we estimate their quality as proxies? While annotation artifacts are one facet that affects proxy-quality \cite{gururangan-etal-2018-annotation, poliak-etal-2018-hypothesis, kaushik-lipton-2018-much, naik-etal-2018-stress, glockner-shwartz-goldberg:2018:Short}, a dataset might simply not have coverage across competencies required for a task. Additionally, it might consist of alternate ``explanations'', features correlated with the task label in the dataset while not being task-relevant, which models can exploit to give the impression of good performance at the task itself.

Two analysis methods have emerged to address this limitation: 1) \textbf{Diagnostic examples}, where a small number of samples in a test set are annotated with linguistic phenomena of interest, and task accuracy is reported on these samples \cite{williams-etal-2018-broad, joshi-etal-2020-taxinli}. However, it is difficult to determine if models perform well on diagnostic examples because they actually learn the linguistic competency, or if they exploit spurious correlations in the data \cite{mccoy-etal-2019-right, gururangan-etal-2018-annotation, poliak-etal-2018-hypothesis}.  2) \textbf{External challenge tests} \cite{naik-etal-2018-stress, isabelle-etal-2017-challenge, glockner-shwartz-goldberg:2018:Short,ravichander-etal-2019-equate, mccoy-etal-2019-right}, where examples are constructed, either through automatic methods or by experts, exercising a specific phenomenon in isolation. However, it is challenging and expensive to build these evaluations, and non-trivial to isolate phenomena \cite{liu2019inoculation}.

 Thus, \textbf{probing} or \emph{diagnostic classification} presents a compelling alternative, wherein learned representations can directly be probed for linguistic properties of interest \cite{ettinger2016probing,belinkov2017neural, adi2017finegrained, tenney2018what,zhang-bowman-2018-language,  warstadt-etal-2019-investigating}.  There has been a variety of research that employs probing to test hypotheses about the mechanisms models used to perform tasks. \newcite{shi2016does} examine learned representations in machine translation for syntactic knowledge. \newcite{vanmassenhove2017investigating} investigate aspect in neural machine translation systems, finding that tense information could be extracted from the encoder, but that part of this information may be lost when decoding. \newcite{conneau-etal-2018-cram} use probing to examine the correlation between linguistic properties and downstream tasks (including MT and NLI). \newcite{hupkes2018visualisation} train a 'diagnostic classifier' to extract information from a sequence of hidden representations in a neural network. If the classifier achieves high accuracy, it is concluded that the network is keeping track of the hypothesized information. \newcite{giulianelli2018hood} use diagnostic classifiers to predict number from the internal states of a language model.  \citet{kim-etal-2019-probing} study what different NLP tasks teach models about function word comprehension.  \newcite{alt-etal-2020-probing} analyze learned representations in relation extraction, through a set of fourteen probing tasks for relevant linguistic properties. \newcite{saleh-etal-2020-probing} examine the representations learned by neural dialog models for insights into what the model learns about engaging in dialog. See the survey by \newcite{belinkov2019analysis} for many more examples. 

Closely related to our work is that of \citet{hewitt-liang-2019-designing}, which studies the role of lexical memorization in probing, and recently the work of \citet{pimentel-etal-2020-information} and \citet{voita-titov-2020-information} who analyze probing from an information-theoretic perspective.  These works join an ongoing debate on the correct way to characterize the expressivity of the probing classifier, with the latter proposing ease of extractability as a criterion for selecting appropriate probes. Our work pursues an orthogonal line of inquiry, demonstrating that relying on diagnostic classifiers to interpret model reasoning for a task suffers from a fundamental limitation: properties may be incidentally encoded even when not required for a task. Thus, our work is also related to a broader investigation of how neural models encode information \cite{tishby2015deep,voita2019bottom}, studying to what extent information encoded in neural representations is indicative of information needed to perform tasks.

\section{Methodology}
In this section we describe our modified probing pipeline (Figure~\ref{ref:probing-exp}), where we construct control datasets, such that a particular linguistic feature is not required in making task judgements.\footnote{While our motivating example of a task is natural language inference, we expect control datasets can be constructed for most text classification tasks with a small finite label space.} Control datasets are based on the intuition that a linguistic feature is not informative for a model to discriminate between classes if the linguistic feature remains constant across classes. For a  task label $T$ and linguistic property $L$, when every example in the control dataset has the same value for $L$, the linguistic property $L$ in isolation is not discriminative of the task label .

To construct control datasets we hold constant the relevant property value across the whole dataset. In practice, the control datasets are constructed from existing large-scale datasets by partitioning them on the value of a linguistic property, such that every example in the sampled dataset has the same value of linguistic property.\footnote{All probing tasks in this work take a sentence representation as input and perform mappings to binary labels \{0, 1\}. } They are designed with the following considerations:
\begin{enumerate}[itemsep=2pt,topsep=2pt,parsep=2pt]
\item 
 The linguistic property of interest is auxiliary to the main task and a function of the input, but not of the task decision.
\item 
Every sample in the training and test sets has the same fixed value of the linguistic property. 
\item 
The training set is large in order to train parameter-rich neural classifiers for the task.
\end{enumerate}

We next describe our main training task, our three auxiliary prediction tasks, and procedures to construct control datasets corresponding to each auxiliary property. Models are trained either on datasets constructed for the main task, or on control datasets, and then probed for the auxiliary property using data from a probing dataset. In this work, we use the experimental settings of \citet{conneau-etal-2018-cram} for both the training task and probing task, due to its popularity as a probing benchmark. However, the conclusions we draw are meant to illustrate the limits and generality of probing as a diagnostic method, rather than discuss the specific experimental settings of \citet{conneau-etal-2018-cram}.

\begin{table}[tb]
\small
\begin{center}
\begin{tabular}{ l l l l } \toprule
Linguistic Control & Property & \# Train & \# Test \\ \midrule
MultiNLI         & -              &  392,702        &  20,000       \\ %
Tense          & Past     &       69,652   &    1678     \\ %
 Subject Number  & Singular    &     102,452     &   2584      \\ %
 Object Number       & Singular &   43,178      & 1060 \\ \bottomrule
\end{tabular}
\caption{Statistics of control datasets partitioned by linguistic property.}
\label{tab:datasetstats}
\end{center}
\end{table}

\begin{table*}
         \centering
  \begin{tabular}{l @{\hskip 0.5em} c c c@{\hskip 0.3em} c c c@{\hskip 0.3em} c c}
  \toprule
    & \multicolumn{2}{c}{Tense} & & \multicolumn{2}{c}{Subject Number}  & & \multicolumn{2}{c}{Object Number}   \\ 
    \cmidrule{2-3} \cmidrule{5-6} \cmidrule{8-9}
    & Dev-ST & Probing & & Dev-SS & Probing & & Dev-SO & Probing \\
  \midrule
  {Majority} & 37.90 & 50.00 && 36.88  & 50.0  & & 39.52   & 50.0 \\ \hline
CBOW-DS & 57.57 & 82.36 && 58.4 & 76.55 && 55.85 & 75.49\\ 
CBOW-PT & 60.31 & 82.2 && 58.2 & 75.69 && 59.15 & 74.38\\ \hline
BiLSTM-Av-DS & 63.53 & 82.93 && 64.24 & 79.53 && 66.23 & 76.11\\ 
BiLSTM-Av-PT & 65.08 & 82.79 && 66.76 & 78.81 && 67.08 & 75.48\\ \hline
BiLSTM-Max-DS & 63.35 & 81.14 && 65.91 & 78.56 && 65.94 & 74.79\\ 
BiLSTM-Max-PT & 64.6 & 81.04 && 66.87 & 79.51 && 66.98 & 72.44\\ \hline
BiLSTM-Last-DS & 61.08 & 80.43 && 64.2 & 81.52 && 62.26 & 72.65\\ 
BiLSTM-Last-PT & 63.89 & 78.44 && 66.18 & 78.9 && 66.04 & 72.82\\ 
  \bottomrule
  \end{tabular}

\caption{Performance comparisons of %
task-controlled (PT) and downsampled models (DS). Dev-ST, Dev-SS and Dev-SO is the MultiNLI development set controlled for tense, subject number and object number, respectively. PT is a model trained on data partitioned by linguistic property---these models should not be able to leverage the linguistic property to perform their training task. DS is models trained on downsampled MNLI data to match the number of instances in partitioned. Majority baseline reflects distribution of main task classes for controlled development sets (Dev-ST, Dev-SS and Dev-SO), or class distribution of auxiliary property for probing datasets. We can observe that models consistently display similar probing accuracies whether the property was needed for the training task or not (Probing). Competitive performance of PT model variants to DS model variants on controlled MNLI development sets (Dev-ST, Dev-SS, Dev-SO) validates the controlled linguistic property is not useful to solve the controlled version of the task.}
\label{tab:ModelPerfFirst}

\end{table*}

\begin{table*}[tb]
         \centering
  \begin{tabular}{l @{\hskip 0.5em} c c c@{\hskip 0.3em} c c c@{\hskip 0.3em} c c}
  \toprule
    & \multicolumn{2}{c}{Tense} & & \multicolumn{2}{c}{Subject Number}  & & \multicolumn{2}{c}{Object Number}   \\ 
    \cmidrule{2-3} \cmidrule{5-6} \cmidrule{8-9}
    & Dev & Probing & & Dev & Probing & & Dev & Probing \\
  \midrule
  Majority & 36.50 & 50.0 &  & 36.50  & 50.0  & & 36.50 & 50.0  \\ \hline
  CBOW-Word  & 62.21 & 83.74 && 62.1 & 76.91 && 61.93 & 75.4\\ 
CBOW-Rand & 56.98 & 60.14 && 56.27 & 67.01 && 56.82 & 64.71\\ \hline 
BiLSTM-Avg-Word & 70.05 & 82.48 && 70.67 & 76.53 && 69.82 & 72.29\\ 
BiLSTM-Avg-Rand  & 63.33 & 61.4 && 64.0 & 67.68 && 63.71 & 63.87\\ \hline 
BiLSTM-Max-Word & 68.67 & 78.34 && 69.19 & 73.96 && 69.12 & 68.53\\ 
BiLSTM-Max-Rand  & 62.78 & 62.89 && 63.29 & 69.51 && 63.28 & 62.84\\ \hline 
BiLSTM-Last-Word & 68.32 & 74.61 && 69.04 & 71.82 && 68.82 & 69.27\\ 
BiLSTM-Last-Rand & 62.14 & 62.96 && 61.88 & 67.45 && 62.29 & 61.32\\  
  \bottomrule
  \end{tabular}

\caption{Performance comparisons of models initialized with pretrained word embeddings (Word) and models with randomly initialized embeddings (Rand) on MNLI Development Set (Dev) and on the probing task (Probing). Embeddings are updated during task-specific training. We can observe that probing performance decreases sharply for all models when word embeddings are randomly initialized, suggesting a considerable component of probing performance comes from pretraining word embeddings rather than what a model learns during the task.}

\label{tab:ModelPerf_embedrand}
\end{table*}

\noindent
\paragraph{Main Task:}
We study the Natural Language Inference (NLI) training task from \citet{conneau-etal-2018-cram} as the main task for training sentence encoders. NLI is a benchmark task for research on natural language understanding \cite{cooper1996using, haghighi2005robust, harabagiu2006methods, dagan2006pascal,giampiccolo2007third, zanzotto2006learning, maccartney2009natural, dagan2010fourth, marelli-etal-2014-sick, williams-etal-2018-broad}. Broadly, the goal of the task is to decide if a given hypothesis can be inferred from a premise in a justifiable manner. Typically, this is framed as the 3-way decision of whether a hypothesis is true given the premise (entailment), false given the premise (contradiction), or whether the truth value cannot be determined (neutral). We use  MultiNLI \cite{williams-etal-2018-broad}, a broad-coverage NLI dataset, to train sentence encoders. %

\noindent
\paragraph{Auxiliary Tasks:} We consider three tasks that probe sentence representations for semantic information from \citet{conneau-etal-2018-cram}, all of which ``require some understanding of what the sentence denotes''. We construct the probing datasets such that lexical items that are associated with the probing task do not occur across the train/dev/test split for the target. This design controls for the effect of memorizing word types associated with target categories \cite{hewitt-liang-2019-designing}. The tasks considered in this study are:
\begin{enumerate}[itemsep=-0.2em,topsep=3pt,parsep=3pt]
\item \textsc{Tense}: Categorize sentences based on the tense of the main verb. 
\item \textsc{Subject Number}: Categorize sentences based on the number of the subject of the main clause.
\item \textsc{Object Number}: Categorize sentences based on number of the direct object of the main clause.
\end{enumerate}

\noindent
\paragraph{Control:}
\label{ref:nlipartitioning} 
 For each auxiliary task, we partition MultiNLI such that premises and hypotheses agree on a single value of the linguistic property. For example, for the auxiliary task \textsc{Tense},  sentences with VBP/VBZ/VBG forms are labeled as present and VBD/VBN as past tense.\footnote{These heuristics are specific to English, as is MultiNLI. We use the Stanford Parser for constituency, POS and dependency parsing \cite{manning-EtAl:2014:P14-5}.} Subsequently, premise-hypothesis pairs where the main verbs in both premise and hypothesis are in past tense are extracted from train/dev sets to form the control datasets for tense. Thus, every sentence in the dataset (both premises and hypotheses), has the same value of the auxiliary property.\footnote{This procedure replicates the original SentEval probing labels \cite{conneau-etal-2018-cram} with 89.37\% accuracy on tense, 87.77\% accuracy on subject number and 88.19\% accuracy on object number. }
 
This procedure results in three control datasets/tasks: MultiNLI-PastTense, MultiNLI-SingularSubject, and MultiNLI-SingularObject. For all three, we fix the value of the linguistic property to the one that results in the maximum number of training instances on partitioning, namely fixing past tense, singular subject number, and singular object number. Descriptive statistics for each dataset appears in Table \ref{tab:datasetstats}. %

\noindent
\paragraph{Models:}  We use CBOW and BiLSTM-based sentence-encoder architectures. The choice of these models is motivated by their demonstrated utility as NLI architectures \cite{williams-etal-2018-broad}, and because their learned representations have been extensively studied for the three linguistic properties used in this work \cite{conneau-etal-2017-supervised}.\footnote{We leave an exploration of recent transformer-based architectures to future work, noting however that this study stands alone as evidence that probing performance does not correlate to task importance.}

\paragraph{1. Majority:} The hypothetical performance of a classifier that always predicts the most frequent label in the test set.
\paragraph{2. CBOW:} A  simple Continuous Bag-Of-Words Model (CBOW). The sentence representation is the sum of word embeddings of constituent words. Word embeddings are finetuned during training.
\paragraph{3. BiLSTM-Last/Avg/Max:} For a sequence of $N$ words in a sentence $s = w_1 ... w_n$, the bidirectional LSTM (BiLSTM; \citet{10.1162/neco.1997.9.8.1735}) computes N vectors extracted from its hidden states $\vec{h}_1, ..., \vec{h}_n$. We produce fixed-length vector representations in three ways: by selecting the last hidden state $h_{n}$ (BiLSTM-Last), by averaging the produced hidden states (BiLSTM-Avg) or by selecting the maximum value for each dimension in the hidden units (BiLSTM-Max).

All models produce separate sentence vectors for the premise and hypothesis.  They are concatenated with their element-wise product and difference \cite{mou-etal-2016-natural}, passed to a \texttt{tanh} layer and then to a 3-way softmax classifier. Models are initialized with 300D GloVe embeddings \cite{pennington2014glove} unless specified otherwise, and implemented in %
Dynet %
\cite{neubig2017dynet}. After the model is trained for the NLI task, the learned sentence vectors for the premise and hypothesis are probed.  The probing classifier is a 1-layer multi-layered perceptron (MLP) with 200 hidden units.

\begin{table*}[tb]
         \centering
  \begin{tabular}{l @{\hskip 0.5em} c c c@{\hskip 0.3em} c c c@{\hskip 0.3em} c c}
  \toprule
    & \multicolumn{2}{c}{Tense} & & \multicolumn{2}{c}{Subject Number}  & & \multicolumn{2}{c}{Object Number}   \\ 
    \cmidrule{2-3} \cmidrule{5-6} \cmidrule{8-9}
    & Dev-ST & Probing & & Dev-SS & Probing & & Dev-SO & Probing \\
  \midrule
  Majority & 37.90 & 50.0 & & 36.88 & 50.0 &  & 39.52 & 50.0 \\ \hline
  CBOW-Rand-DS & 49.88 & 61.33 && 51.04 & 67.32 && 49.25 & 63.63\\ 
CBOW-Rand-PT & 53.28 & 61.37 && 50.97 & 67.02 && 52.45 & 63.84\\ \hline 
BiLSTM-Avg-Rand-DS & 57.21 & 63.75 && 60.76 & 68.5 && 59.53 & 63.89\\ 
BiLSTM-Avg-Rand-PT & 60.91 & 63.07 && 61.18 & 69.12 && 60.57 & 63.77\\ \hline 
BiLSTM-Max-Rand-DS & 59.18 & 61.05 && 61.8 & 70.32 && 60.57 & 64.68\\ 
BiLSTM-Max-Rand-PT & 60.55 & 61.53 && 63.78 & 70.6 && 63.49 & 64.26\\ \hline 
BiLSTM-Last-Rand-DS & 56.73 & 63.88 && 58.82 & 69.09 && 56.79 & 63.86\\ 
BiLSTM-Last-Rand-PT & 57.39 & 62.88 && 61.88 & 68.8 && 60.75 & 61.96\\  
  \bottomrule
  \end{tabular}

\caption{Performance of task-controlled (PT) and downsampled models (DS), when word embeddings are trained from scratch. (Rand) indicates the model is initialized with random embeddings, rather than pretrained embeddings. Dev-ST, Dev-SS and Dev-SO is the MultiNLI development set controlled for tense, subject number and object number, respectively. We observe that when the training task is isolated in this way, for all models probing performance is similar whether a linguistic property is necessary for the task or not (Probing). }

\label{tab:ModelPerf}
\end{table*}

\section{Probing the Probing Paradigm}
\label{ref:probingprobingsection}

\subsection{Probing with Linguistic Controls}
\label{ref:incidentalNLI}
As a first step, we ask the question: to what extent is the information encoded in learned representations, as reflected in probing accuracies, driven by information that is useful for the training task?
We construct multiple versions of the task (both training and development sets) where the entailment decision is independent of the given linguistic property, through careful partitioning as described in \S \ref{ref:nlipartitioning}. To control for the effect of training data size, we downsample MultiNLI training data to match the number of samples in each partitioned version of the task. These results are in Table \ref{tab:ModelPerfFirst}.

Strikingly, we observe that even when models are trained on tasks that \emph{do not require the linguistic property at all} for the main task (rows with PT in Table \ref{tab:ModelPerfFirst}), probing classifiers still exhibit high accuracy  (sometimes up to $\sim$80\%). Probing data is split lexically by target across partitions, and thus lexical memorization \cite{hewitt-liang-2019-designing} cannot explain why these properties are encoded in the sentence representations. Across models,  on the version of the task where a particular linguistic property is not needed, classifiers trained on data that does not require that property perform comparably to classifiers trained on MultiNLI training data (DS vs PT models, on Dev-ST, Dev-SS, and Dev-SO).

\subsection{Effect of Word Embeddings}
\label{ref:effectwordembedding}
A potential explanation lies in our definition of a ``task''. Previous work directly probes models trained for a target task such as NLI. However, when models are initialized with pre-trained word embeddings, the conflated results of two tasks are being probed --  the main training task of interest, and the task that was used to train the word embeddings. Both tasks may contribute to the encoding of information in the learned representation, and it is unclear to what extent they interact. Previous work has noted the considerable amount of information present in word embeddings, and proposed methods to measure this effect, such as comparing with bag-of-word baselines or random encoders \cite{wieting2018no}. However, these methods fail to isolate the contribution of the training task.

To study this, we compare models initialized with pre-trained word embeddings \cite{pennington2014glove} and then trained for the main task, to models initialized with random word embeddings and then updated during the main task. These results are presented in Table~\ref{tab:ModelPerf_embedrand}. We observe that probing accuracies drop across linguistic properties in this setting (compare rows with Word and Rand in the table), indicating that models with randomly initialized embeddings generate representations that contain less linguistic information than the models with pretrained embeddings. This result calls into question how to interpret the contribution of the main task to the encoding of a linguistic property, when the representation has already been initialized with pre-trained word embeddings. The word embeddings could themselves encode a significant amount of linguistic information, or the main task might contribute to encoding information in a way already largely captured by word embeddings.

\subsection{How do models encode linguistic properties?}
\label{ref:isolae}

When we isolate the effect of the main task with randomly initialized word embeddings, are properties not predictive of the main task judgement \emph{still} being encoded? To study this, we revisit our linguistic control tasks but train all models with randomly initialized word embeddings. We also train comparable models on downsampled MultiNLI training data. These results can be found in Table 4. We observe that even in the setting with randomly initialized word embeddings, these properties are still encoded to a similar extent (and above the majority baseline) in the downsampled and control versions of their task.

\section{A Synthetic Experiment: Analyzing Encoding Dynamics}
\label{ref:adversarial}
We have demonstrated that models encode properties even when they are not required for the main task. Thus, probing accuracy cannot be considered indicative of competencies any given model relies on. What circumstances could lead to models encoding properties incidentally? Can we determine when a linguistic property is not needed by a model for a task? To study this, we build carefully controlled synthetic tests, each capturing a kind of noise that could arise in datasets.  

\subsection{Synthetic Task}
\label{ref:synthetictasks}
We consider a task where the premise \emph{P} and hypothesis \emph{H} are strings from $S = \{(a|b)(a|b|c)^{*}\}$ of maximum length 30, and the hypothesis \emph{H} is said to be entailed by the premise \emph{P} if it begins with the same letter \emph{a} or \emph{b},\footnote{A task with a similar objective was used by \citet{belinkov-etal-2019-dont} to demonstrate unlearning bias in datasets. The task is equivalent to XOR, which is learnable by an MLP.} for example: 
\begin{table}[H]
\setlength\tabcolsep{2pt}
\begin{tabular}{p{.5\columnwidth} p{.5\columnwidth}}
(a, ab) $\rightarrow$ Entailed & (a, ba) $\rightarrow$ Not Entailed\\
(b, ba) $\rightarrow$ Entailed & (b, ab) $\rightarrow$ Not Entailed\\
(b, bc) $\rightarrow$ Entailed & (b, acb) $\rightarrow$ Not Entailed\\
\end{tabular}
\end{table}

Consider the auxiliary task of predicting whether a sentence contains the character \emph{c} from a representation, analogous to probing for a task-irrelevant property. We sample premises/hypotheses from a set of strings $S' = (a|b)^{*}$ of maximum length 30, and simulate four kinds of correlations that could occur in a dataset by inserting \emph{c} at a random position in the string after the first character:\footnote{We additionally explore the utility of adversarial learning, as a potential approach to identifying properties required by a model to perform a task, by suppressing a property and measuring task performance (Appendix. \ref{ref:adv-framework}). We find in our exploration that adversarial approaches are not completely successful at suppressing the linguistic property under consideration, though capacity of the adversary could play a role.} %
\begin{enumerate}[itemsep=1pt,topsep=3pt,parsep=3pt]
    \item \textsc{Noise} : The property could be distributed as noise in the training data. To simulate this, we insert \emph{c} into 50\% of randomly sampled premise and hypothesis strings.
    \item \textsc{Uncorrelated} : The property could be unrelated to the task decision, but correlated to some other property in the data. To simulate this, we insert \emph{c} to premises beginning with \emph{a}.
    \item \textsc{Partial}: The property could provide a partial explanation for the main task decision. To simulate this, we insert \emph{c} to premise and hypothesis strings beginning with \emph{a}.\footnote{Models can use either the presence of \emph{c}, or the first character of the strings being \emph{a} to make their prediction, but they must use whether the first character of the strings is \emph{b}.}
    \item  \textsc{Full}: The property provides a complete alternate explanation for the main task decision. We insert \emph{c} to premise and hypothesis strings whenever the hypothesis is entailed.
\end{enumerate}
Descriptive statistics of all datasets are in Table 6.

\begin{table}[tb]
\begin{center}
\begin{tabular}{ | l l l l |} \toprule
Dataset & \# Train & \# Dev & \# Test \\ \midrule
\textsc{Noise}         & 20000              &  5000        &  5000       \\ 
\textsc{Uncorrelated}          & 20000              &  5000        &  5000      \\ 
\textsc{Partial}         & 20000              &  5000        &  5000     \\ 
\textsc{Full}          & 20000              &  5000        &  5000   \\ \midrule
\textsc{Probe}              & 23732              &  5000        &  5000  \\ \bottomrule
\end{tabular}
\caption{Number of train/dev/test examples in constructed synthetic datasets.}
\label{tab:syntheticdatasetstats}
\end{center}
\caption{Descriptive statistics for \textsc{Noise}, \textsc{Uncorrelated}, \textsc{Partial} and \textsc{Full}  synthetic datasets, as well as the dataset used to train the probing classifier(\textsc{Probe}). We ensure that datasets do not have any data leakage in the form of strings appearing across train/dev/test splits, or across task and probing splits in either the main task or the probing dataset.}
\end{table}

\begin{figure*}[tb]

\centering
\begin{subfigure}{\textwidth}
\includegraphics[width=.24\textwidth]{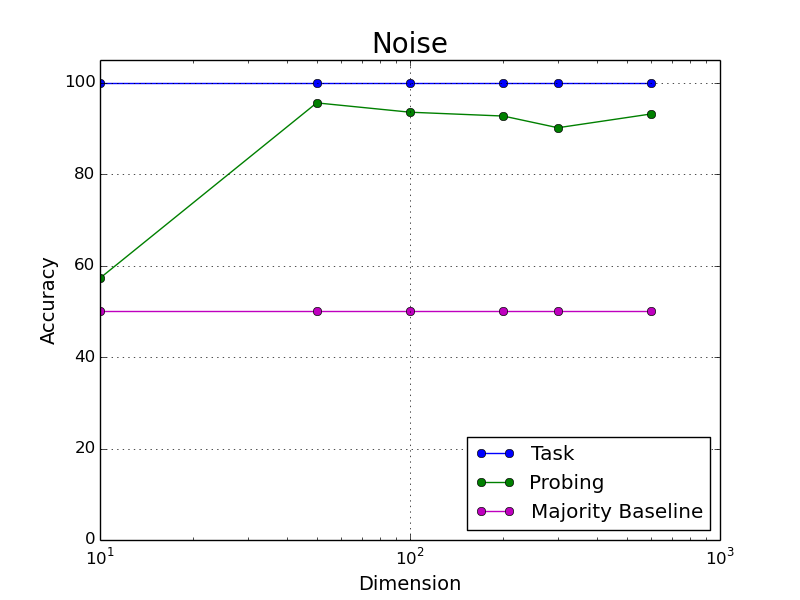}
\hfill
\includegraphics[width=.24\textwidth]{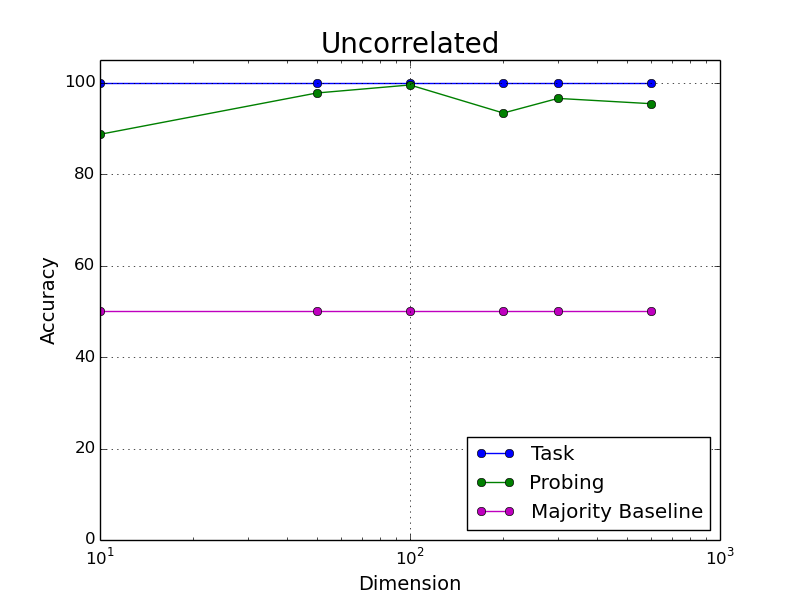}
\hfill
\includegraphics[width=.24\textwidth]{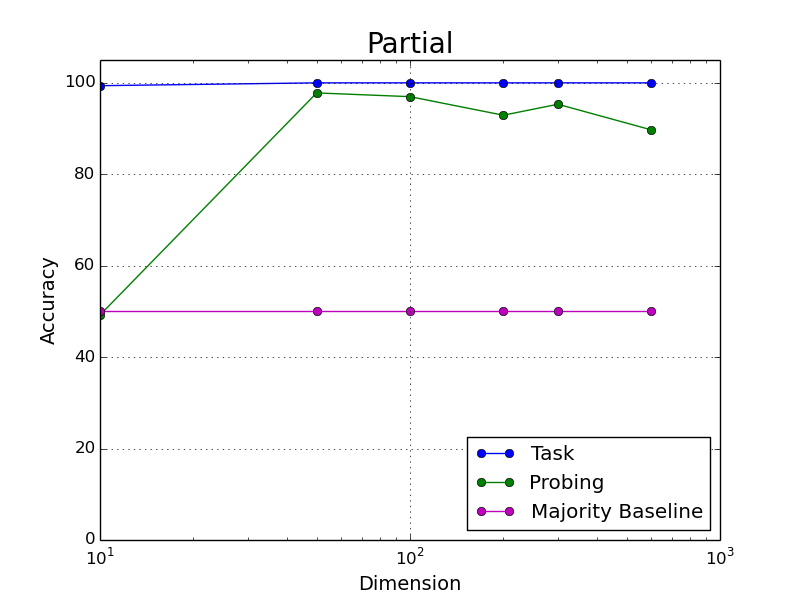}
\hfill
\includegraphics[width=.24\textwidth]{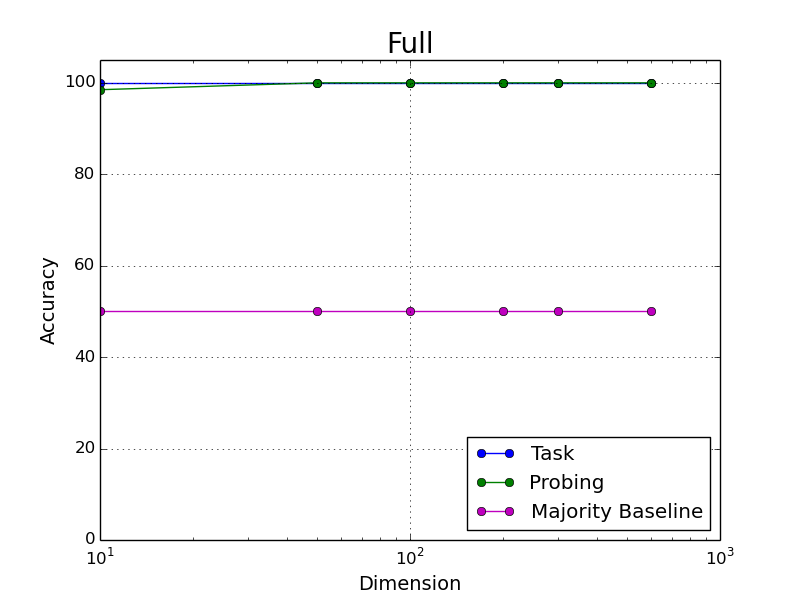}
\caption{Main Task and Probing Accuracy as a function of capacity of sentence representation (\# units).}
\label{ref:repn-capacity-fig}
\end{subfigure}

\centering
\begin{subfigure}{\textwidth}

\includegraphics[width=0.24\textwidth]{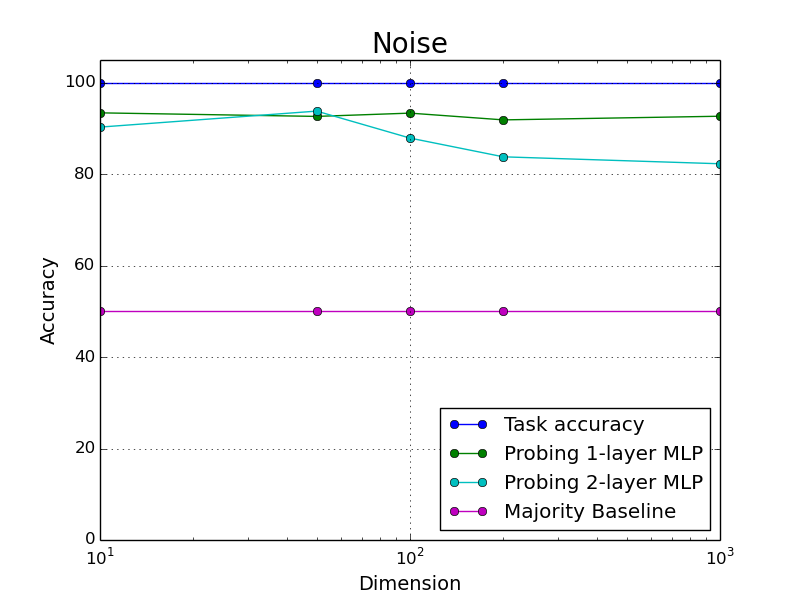}
\hfill
\includegraphics[width=0.24\textwidth]{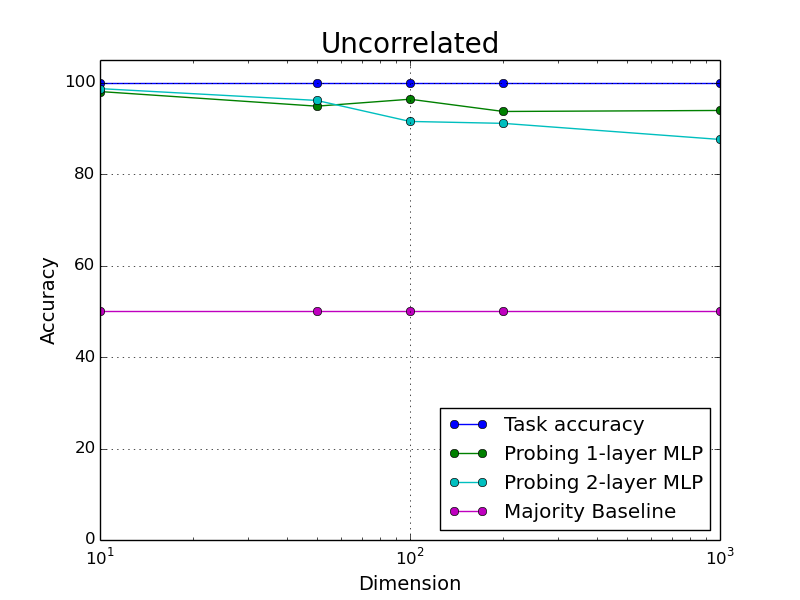}
\hfill
\includegraphics[width=0.24\textwidth]{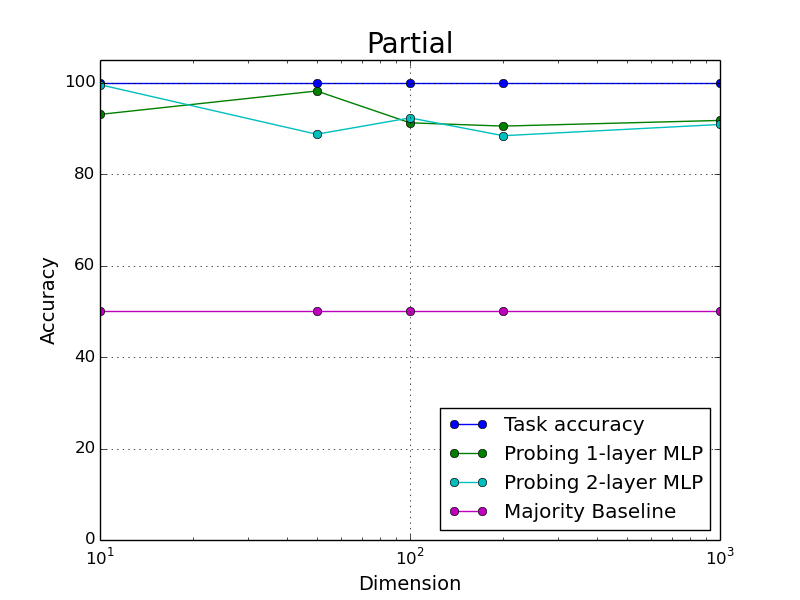}
\hfill
\includegraphics[width=0.24\textwidth]{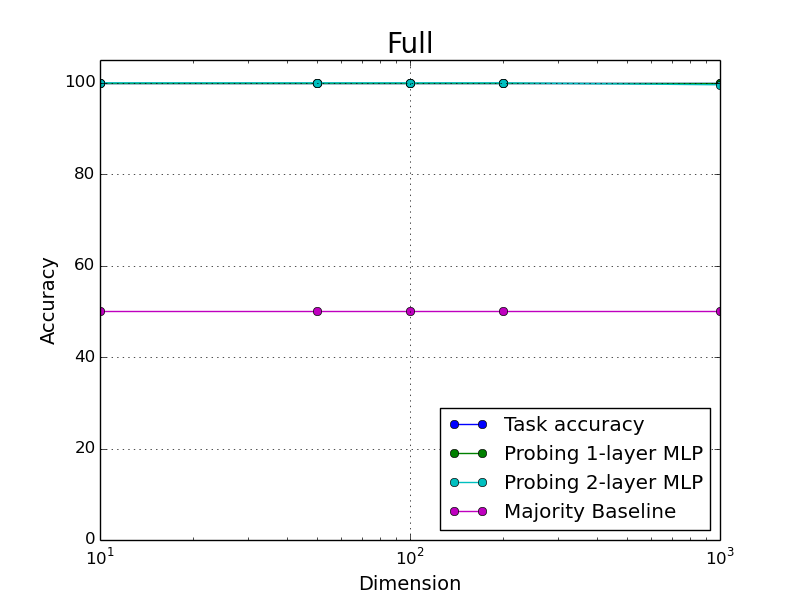}

\caption{Main Task and Probing Accuracy as a function of capacity of probing classifier (\# units).}
\label{ref:prob-capacity-fig}
\end{subfigure}

\caption{Task and probing accuracy of (BiLSTM, last) on Noise, Uncorrelated, Partial and Full synthetic datasets.}

\end{figure*}

\subsection{Results}

Figure \ref{ref:repn-capacity-fig} presents the performance of the model and the probe on the four test sets. We observe that we are able to train a classifier to predict the presence of \emph{c} considerably above chance-level in all four cases. This is notable, considering that even when the property is distributed as random noise (\textsc{Noise}) uncorrelated with the actual task, the model encodes it. This simple synthetic task suggests that models learn to encode linguistic properties \emph{incidentally}, implying it is a mistake to rely on the accuracy of probes to measure what information the model relies upon to solve a task. We further discuss the role of representation capacity and probing classifier expressivity:

\noindent
\paragraph{Representation size:}  Lower-capacity models may encode task-specific information at the expense of irrelevant properties. To examine this, we train the BiLSTM architecture with hidden size 10, 50, 100, 200, 300 and 600 units, and train the probing classifier on the auxiliary task. These results are reported in Figure \ref{ref:repn-capacity-fig}. We observe that while the main task accuracy remains consistent across choice of dimension,  probing accuracy decreases for models with lower capacity across categories. This suggests that the capacity of the representation may play a role in which information it encodes, with lower capacity models being less prone to incidentally encoding irrelevant information.

\paragraph{Probing classifier capacity:}
We examine whether probing classifier capacity is a factor in the incidental encoding of linguistic properties. A more complex probing classifier may be more effective at extracting linguistic properties from representations. We experiment with probing classifiers utilizing 1-layer and 2-layer MLP's of dimensions \{10, 50, 100, 200, 1000\}. The results are shown in Figure~\ref{ref:prob-capacity-fig}. We find that a higher-capacity probing classifier does not necessarily imply higher probing accuracy. Further, in all the settings of probing classifier capacity we study, we are able to perform the auxiliary task considerably above chance accuracy, even when the property is distributed as random noise.

\section{Discussion}
\label{ref:discussion}

We briefly discuss our findings, with the goal of providing considerations for deciding which inferences can be drawn from a probing study, and highlighting avenues for future research. \\

\noindent
\textbf{Linguistic properties can be incidentally encoded}: Probing only indicates that some property correlated with a linguistic property of interest is encoded in the sentence representation --- but we speculate that it cannot isolate what that property might be, whether the correlation is meaningful, or how many such properties exist. As shown in the controlled synthetic tests, even if a particular property is not needed for a task, the information can be extracted from the representation with high accuracy. Thus, probing cannot determine if the property is actually needed to do a task, and should not be used to pinpoint the information a model is relying upon. A negative result here can be more meaningful than a positive one. Adversarially suppressing the property may help determine if an alternate explanation is readily available to the model, with an appropriate choice of probing classifier. In this case, if the model maintains task accuracy while suppressing the information, one can conclude the property is not needed by the model for the task, but its failure to do so is not indicative of property importance. Causal alternatives to probing classifiers that intervene in model representations to examine effects on prediction also present another promising direction for future work ~\cite{giulianelli2018hood,bau2018gan,vig2020causal}.\\

\noindent
\textbf{Careful controls and baselines}:
We emphasize the need for probing work to establish careful controls and baselines when reporting experimental results. When probing accuracy for a linguistic competence is high, it may not be directly attributable to the training task. In this work we identify two confounds: incidental encoding and interaction between training tasks. Perhaps future work will determine causes of incidental encoding and identify further baselines and controls that allow reliable conclusions to be drawn from probing studies.  \\

\noindent
\textbf{Lack of gold-standard data of task requirements}:
While prior work has discussed the different linguistic competencies that might be needed for a task based on the results of probing studies, these claims are inherently hard to reliably quantify given that the exact linguistic competencies, as well as the extent to which they are required, are difficult to isolate for most real-world datasets. Controlled test cases (such as those in \S \ref{ref:synthetictasks}) are effective  as basic sanity checks for claims based on diagnostic classification, and provide insight into encoding dynamics in sentence representations. \\

\noindent
\textbf{Datasets are proxies for tasks, and proxies are imperfect reflections}:
Finally, we speculate that while datasets are used as proxies for tasks, they might not reflect the full complexity of the task. Aside from having dataset-specific idiosyncrasies in the form of unwanted biases and correlations, they might also not require the full range of competencies that we expect models to need to succeed on the task. Future work should refine or move beyond the probing paradigm to carefully identify what the competencies reflected in any dataset are, and how representative they are of overall task requirements.\\

\noindent
\textbf{What probes are good for}:
This work explores only the implications of probing as a diagnostic tool for pinpointing the information models use to do a task. However, when sentence representations are used subsequently downstream (after being trained on the main task), probing can give insight into what information is encoded in the model (irrespective of how that encoding came to be). Future work could include exploring the connection between information encoded in the representation and whether models successfully learn to use them in downstream tasks.

\section{Conclusion}
The probing paradigm has evoked considerable interest as a useful tool for model interpretability. In this work, we examine the utility of probing for providing insights into what information models rely on to do tasks, and requirements for tasks themselves. We identify several considerations when probing sentence representations, most strikingly that linguistic properties can be incidentally encoded even when not needed for a main task. This line of questioning highlights several fruitful areas for future research: how to successfully identify the set of linguistic competencies necessary for a dataset, and consequently how well any dataset meets task requirements, how to reliably identify the exact information models rely upon to make predictions, and how to draw connections between information encoded by a model and used by a model downstream.

\section*{Acknowledgements}
This research was supported in part by grants from
the National Science Foundation Secure and Trustworthy Computing program (CNS-1330596, CNS15-13957, CNS-1801316, CNS-1914486) and a
DARPA Brandeis grant (FA8750-15-2-0277). The
views and conclusions contained herein are those
of the authors and should not be interpreted as
necessarily representing the official policies or endorsements, either expressed or implied, of the
NSF, DARPA, or the US Government.
This research was also supported by the ISRAEL SCIENCE FOUNDATION (grant No. 448/20).
Y.B. was also supported by the Harvard Mind, Brain, and Behavior Initiative. The authors would like to extend
special gratitude to Carolyn Rose and Aakanksha
Naik, for insightful discussions related to this work.
The authors are also grateful to Yanai Elazar, Lucio Dery, John Hewitt, Paul
Michel, Shruti Rijhwani and Siddharth Dalmia for
reviews while drafting this paper, and to Marco
Baroni for answering questions about the SentEval
probing tasks.
\bibliographystyle{acl_natbib}
\bibliography{anthology,eacl2021}
\appendix
\clearpage

\section{Adversarial Learning Framework}  
\label{ref:adv-framework}
We explore an adversarial framework, as a potential approach to identifying incidentally-encoded properties. We study the utility of this framework within the controlled setting of the synthetic task described in Section \ref{ref:adversarial}, where a hypothesis H is entailed by a premise P, if they both begin with the same letter 'a' or 'b'. 
\begin{table}[tb]
\small
\begin{center}
\begin{tabular}{ | l l l l |} \toprule
Dataset & \# Train & \# Dev & \# Test \\ \midrule
\textsc{Noise}         & 20000              &  5000        &  5000       \\ 
\textsc{Uncorrelated}          & 20000              &  5000        &  5000      \\ 
\textsc{Partial}         & 20000              &  5000        &  5000     \\ 
\textsc{Full}          & 20000              &  5000        &  5000   \\ \midrule
\textsc{Attacker}              & 23732              &  5000        &  5000  \\ \bottomrule
\end{tabular}
\end{center}
\caption{Number of train/dev/test examples in constructed synthetic datasets.}
\label{tab:syntheticdatasetstats-appendixfull}
\end{table}

 We train an adversarial classifier to suppress task-irrelevant information, in this case the presence of `c'. The goal is to analyze whether adversarial learning can help a model ignore this information while maintaining task performance. If the model succeeds, it indicates the model does not need the linguistic property for the task. 
Table~\ref{tab:syntheticdatasetstats-appendixfull} provides descriptive statistics for Noise, Uncorrelated, Partial and Full synthetic datasets, as well as the probing dataset used to train the external attack classifier. We ensure that datasets do not have any data leakage in the form of strings appearing across train/dev/test splits, or across task and probing splits in either the main task or the external held-out attacker dataset.

\begin{figure}[tb]
  \centering
  \begin{subfigure}[b]{0.5\columnwidth}
    \includegraphics[width=\columnwidth]{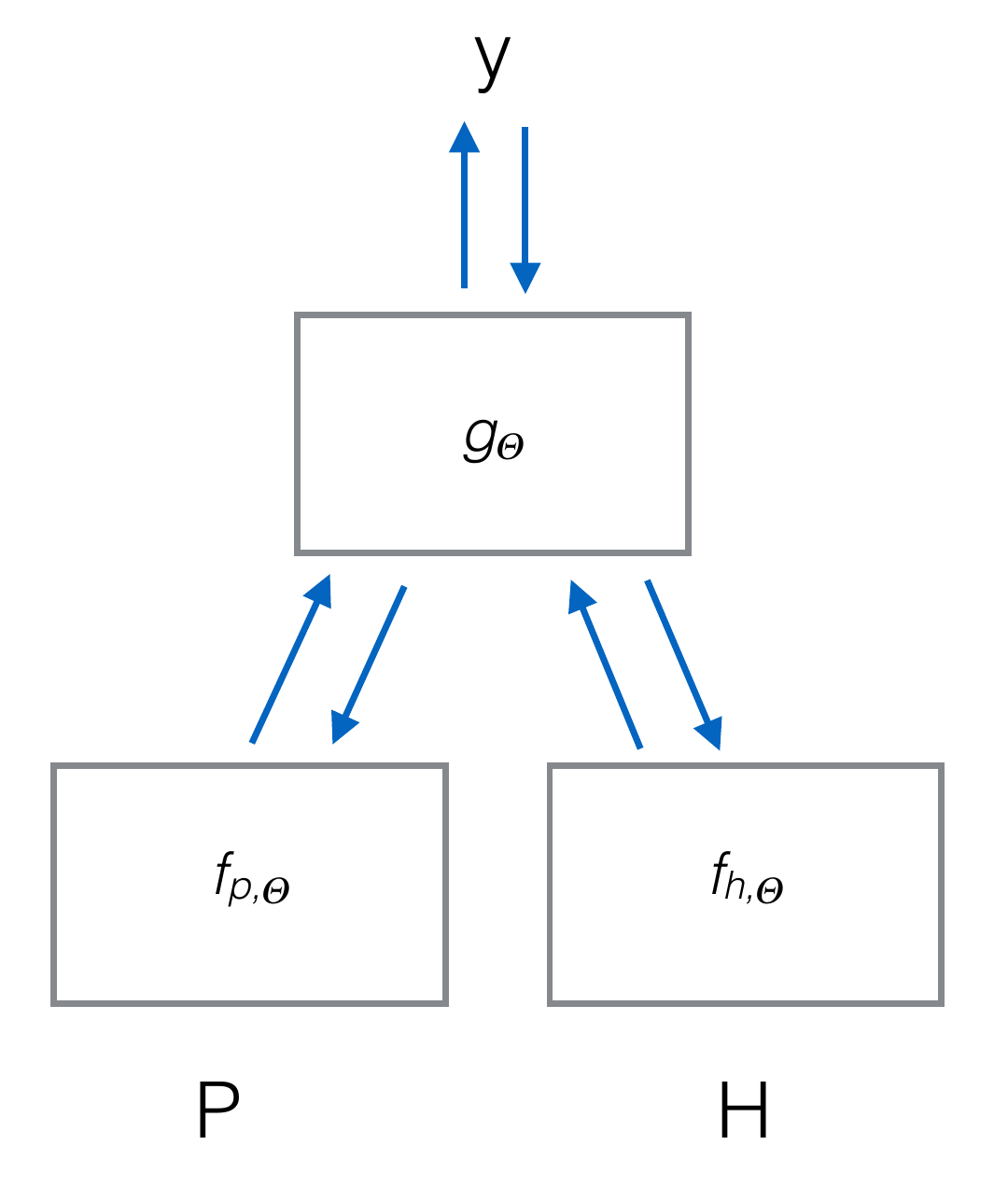}
    \caption{Baseline}
  \end{subfigure}
  \hfill
  \begin{subfigure}[b]{0.65\columnwidth}
    \includegraphics[width=\columnwidth]{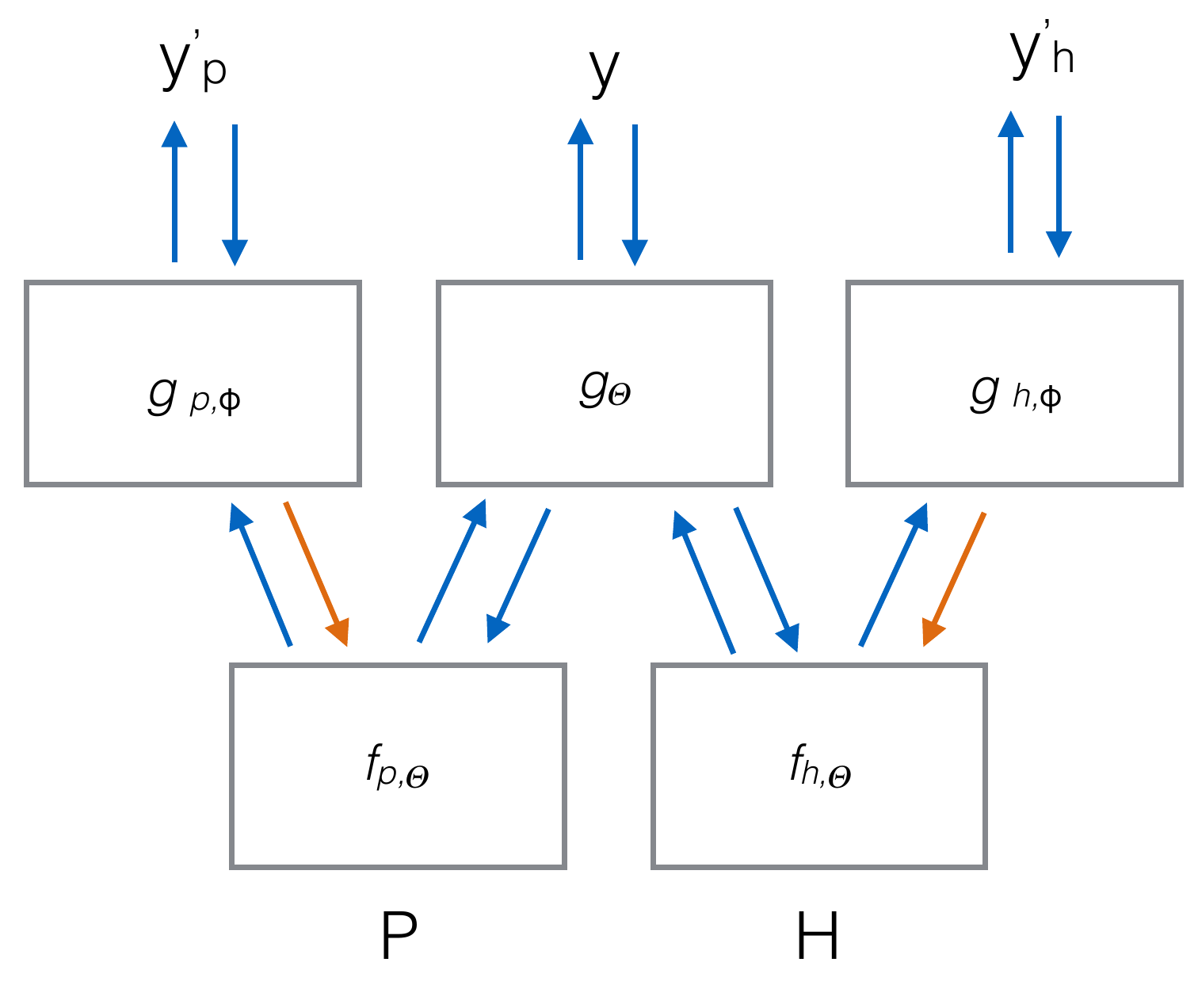}
    \caption{Adversarial removal.}
  \end{subfigure}
  \caption{Illustration of (a) The baseline NLI task architecture, and (b) Adversarial removal of linguistic properties from the representations. Arrows represent direction of propagation of inputs in the forward pass and gradients in backpropagation. Blue and orange arrows correspond to the gradient being preserved and reversed respectively. }
 \label{ref:adv-architecture}
\end{figure}

\begin{table*}[tb]
         \centering
         \small
  \begin{footnotesize}
  \begin{tabular}{l  c c c  c c c   c c c  c c c}
  \toprule
  &  \multicolumn{3}{c}{Noise}   & \multicolumn{3}{c}{Uncorrelated} &  \multicolumn{3}{c}{Partial} &  \multicolumn{3}{c}{Full}  \\ 
    \cmidrule(lr){2-4} \cmidrule(lr){5-7} \cmidrule(lr){8-10} \cmidrule(lr){11-13}
    & Dev & Adv. & Attack. & Dev & Adv. & Attack. &  Dev & Adv. & Attack.&  Dev & Adv. & Attack. \\
  \midrule
Majority  & 50.4 & 51.2 & 50.2 & 50.94 & 74.31 & 50.2 & 50.62 & 99.82 & 50.2 & 55.34 & 55.34 & 50.2  \\
$\lambda$=0.0 & 100.0 & - & 90.3 & 100.0 & - & 93.6 & 100.0 & - & 91.08 & 100.0 & - & 100.0  \\
$\lambda$=0.5 & 100.0 & 47.81 & 95.3 & 100.0 & 70.36 & 62.26 & 100.0 & 99.31 & 80.48 & 100.0 & 51.23 & 93.42  \\
$\lambda$=1.0 & 100.0 & 49.43 & 94.5 & 100.0 & 71.28 & 74.1 & 100.0 & 99.79 & 68.8 & 100.0 & 52.37 & 92.58   \\
$\lambda$=1.5 & 100.0 & 42.7 & 100.0 & 100.0 & 71.54 & 99.1 & 97.98 & 99.79 & 82.32 & 100.0 & 49.8 & 97.58 \\
$\lambda$=2.0 & 100.0 & 46.19 & 99.36 & 100.0 & 70.62 & 99.98 & 100.0 & 94.83 & 91.12 & 100.0 & 40.94 & 94.64 \\
$\lambda$=3.0 & 100.0 & 46.98 & 94.64 & 100.0 & 70.92 & 99.8 & 99.26 & 99.19 & 79.66 & 100.0 & 53.08 & 87.0 \\
$\lambda$=5.0 & 99.98 & 38.87 & 96.92 & 99.94 & 71.0 & 86.6 & 100.0 & 98.73 & 100.0 & 100.0 & 51.32 & 98.74 \\
  \bottomrule
  \end{tabular}
\end{footnotesize}
\caption{Adversarial performance on synthetic tasks: noise, uncorrelated, partial, full. Dev is accuracy of model on task, Adv. is accuracy of the adversarial classifier, Atttack. is accuracy of attacker classifier on held-out data.}
\label{tab:ModelPerf_adv_last_main}
\end{table*}

\begin{figure*}[tb]

\centering
\begin{subfigure}{\textwidth}

\includegraphics[width=.24\textwidth]{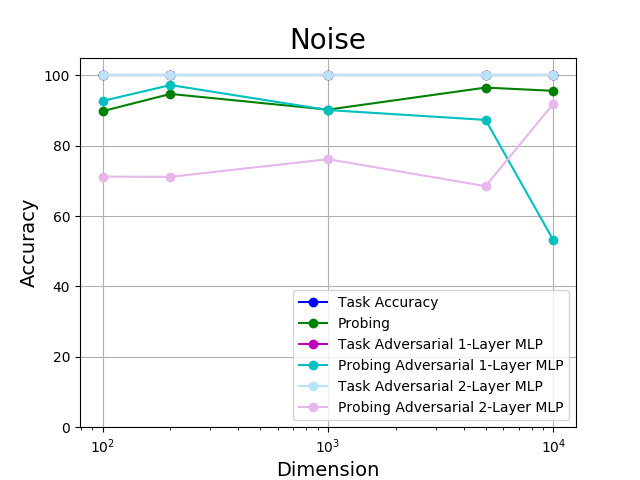}
  \includegraphics[width=.24\textwidth]{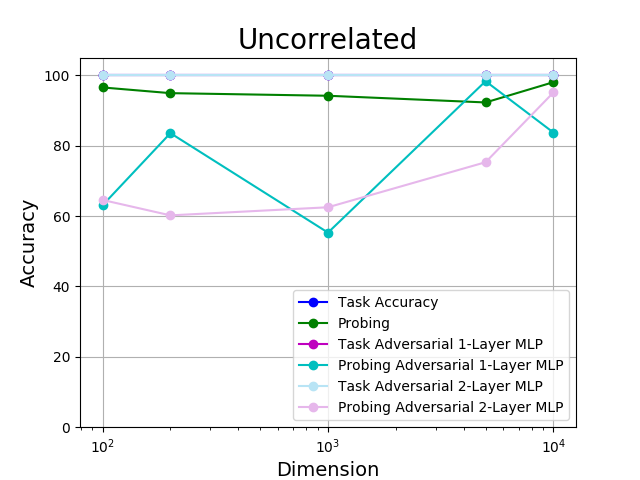}
  \includegraphics[width=.24\textwidth]{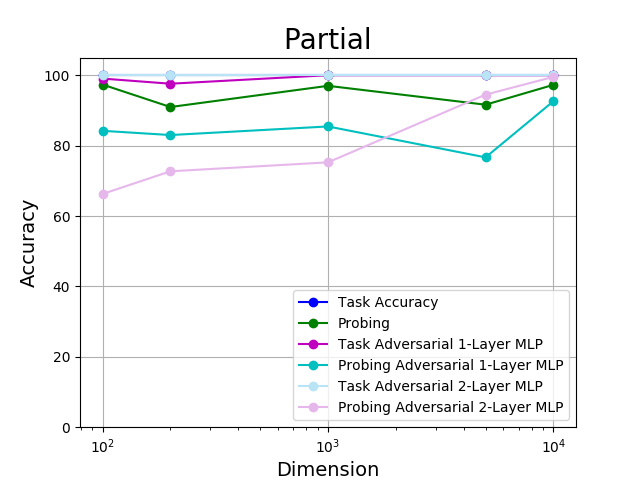}
  \includegraphics[width=.24\textwidth]{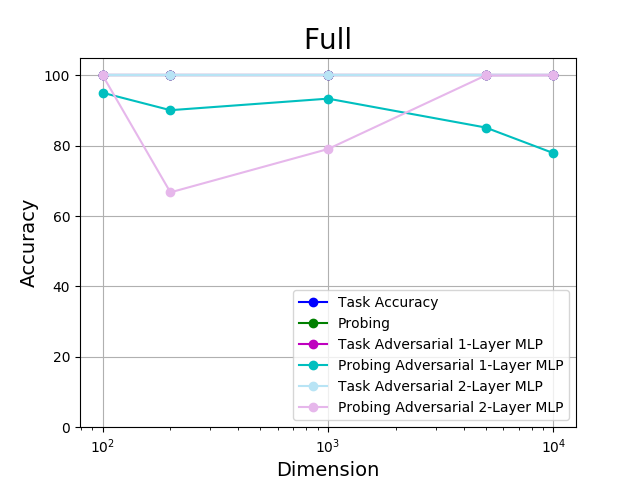}

\end{subfigure}

\caption{Main Task and Attacker Accuracy as a function of capacity of adversarial classifier for $\lambda=0.0$ and $\lambda=1.0$.}
\label{ref:adv-capacity-fig}
\end{figure*}

We follow the adversarial learning framework illustrated in Figure \ref{ref:adv-architecture}.
 In this setup, we have premise-hypothesis pairs $\langle p_{1},h_{1} \rangle ... \langle p_{n},h_{n} \rangle$ and entailment labels $y_{1} ... y_{n}$, as well as labels for linguistic properties in each premise--hypothesis pair $\langle z_{p,1},z_{h,1} \rangle  ... \langle z_{p, n}, z_{h, n} \rangle$. We would like to train sentence encoders f($p_{i}, \theta $) and f($h_{i}, \theta$) and a classification layer $g_{\theta}$ such that $y_{i}= g_{\theta}$ (f($p_{i}, \theta $), f($h_{i}, \theta$)), in a way that does not use $\langle z_{p,i},z_{h,i} \rangle$. We do this by incorporating an \emph{adversarial} classification layer $g_{\phi}$ such that $\langle z_{p,i},z_{h,i} \rangle$ = $\langle g_{\phi}$(f($p_{i}, \theta$)), $g_{\phi}$(f($h_{i}, \theta$)$\rangle$ \cite{goodfellow2014generative, ganin2015unsupervised}. Following \citet{elazar2018adversarial}, we also have an external `attacker' classifier $\phi'$ to predict $z_{p,i}$ and $z_{h,i}$ from the learned sentence representation.\footnote{We train the attacker on a held-out dataset with the linguistic property distributed as random noise (Table~\ref{tab:syntheticdatasetstats-appendixfull}). We also ensure all examples in the attacker data are unseen in the main task, to prevent data leakage. }  A similar setup has been used by \citet{belinkov-etal-2019-adversarial} to remove hypothesis-only biases from NLI models. 

In training the adversarial classifier is trained to predict $z$ from the sentence representations $f_{\theta}(p_{i}, h_{i})$, and the sentence encoder $f$ is trained to make the adversarial classifier unsuccessful at doing so. This is operationalized through the following training objectives optimized jointly:
\vspace{-2mm}
\begin{multline}
\arg\min_{\phi}L(g_{\phi}( f(p_{i}, \theta ),z_{p, i}))\\ + L(g_{\phi}(f(h_{i}, \theta), z_{h, i})) \end{multline}

\vspace{-10mm}

\begin{multline}
\arg\min_{f,\theta} L(g_{\theta}(f_{\theta}(p_i, h_i)), y_i)- (L\\(g_{\phi}( f(p_{i}, \theta ),z_{p, i}) + L(g_{\phi}(f(h_{i}, \theta), z_{h, i})))
\end{multline}
where L is cross-entropy loss. The optimization is implemented through a Gradient Reversal Layer \cite{ganin2015unsupervised} $g_{\lambda}$ which is placed between the sentence encoder and the adversarial classifier. It acts as an identity function in the forward pass, but during backpropogation scales the gradients by a factor $-\lambda$ \footnote{$\lambda$ controls the extent to which we try to suppress the property.}, resulting in the objective:
\vspace{-4mm}
\begin{multline}
\arg\min_{f,\theta}L(g_{\theta}(f_{\theta}(p_i, h_i)), y_i) + L(g_{\phi}(g_{\lambda}(f(p_i\\,\theta))), z_{p,i}) + L(g_{\phi}(g_{\lambda}(f(h_i,\theta))), z_{h,i})  
\end{multline}
\vspace{-6mm}

\paragraph{Implementation details}: We implemented the adversarial model using the Dynet framework \cite{neubig2017dynet}, with a BiLSTM architecture of hidden dimension 200 units. Fixed length vector representations are constructed using the last hidden state and the model is trained for 10 epochs using early stopping. The attacker classifier is a 1-layer MLP with hidden size of 200 dimensions.

\paragraph{Results}
Table~\ref{tab:ModelPerf_adv_last_main} reports the performance of the adversarial and attacker classifiers on the four test sets. We observe that information about irrelevant properties can be extracted by attackers even under adversarial suppression, consistent with the findings of \citet{elazar2018adversarial} and \citet{belinkov-etal-2019-adversarial}. In the case of random noise, we do not find any setting of adversary weight $\lambda$ that suppresses the attribute. We further explore issues of  adversarial classifier strength:

\paragraph{Adversarial classifier capacity:}  A more powerful adversarial classifier may be more effective at suppressing task-irrelevant information. To examine this, we fix the attacker classifier and experiment with adversarial classifiers with 1-layer and 2-layer MLP probes and dimensions 100, 200, 1000, 5000 and 10000 units, as reported in Figure~\ref{ref:adv-capacity-fig}. We find that varying the capacity of the adversarial classifier can decrease the attacker accuracy, though the choice of capacity depends on the setup used. 

\paragraph{Considerations:} 1) In synthetic tests, the main task function is learnable by a neural network. However, in practice for most NLP datasets this might not be true, making it difficult for models to reach comparable task performance while suppressing correlated linguistic properties. 2) Information might be encoded, but may still not be  recoverable by the choice of probing classifier\footnote{All claims related to probing task accuracy, as in most prior work, are with respect to the probing classifier used.}. 3) Adversarial learning does not remove all information from the representation \cite{elazar2018adversarial}. 4) If comparable task accuracy can't be reached, one cannot conclude a property is not relevant.\footnote{This could be because the main task might be more complex to learn or unlearnable, or multiple alternate confounds could be present in data which are not representative of the decision-making needed for the main task.}

\end{document}